\documentclass[11pt,twocolumn,letterpaper]{article}
\usepackage[pagenumbers]{wacv} % To force page numbers, e.g. for an arXiv version

\usepackage{amsmath}
\usepackage{amssymb}
\usepackage{booktabs}
\usepackage{adjustbox}
\usepackage[hidelinks]{hyperref}
\usepackage{balance}
\usepackage[table]{xcolor}

\begin{document}

\title{\textbf{SilVar}: Speech-Driven Multimodal Model for Reasoning Visual Question Answering and Object Localization}

\author{Tan-Hanh Pham$^{1}$, Hoang-Nam Le$^{2}$, Phu-Vinh Nguyen$^{3}$, Chris Ngo$^{4}$, and Truong Son Hy$^{5}$\\
\hphantom{text}\\
$^1$Florida Institute of Technology, USA\\
$^2$FPT University, Vietnam, $^3$Uppsala University, Sweden, \\
$^4$Knovel Engineering Lab, Singapore, $^5$University of Alabama at Birmingham, USA\\ 
\hphantom{text}\\
{\tt\small hanhpt.phamtan@gmail.com}
}

\maketitle

%%%%%%%%% ABSTRACT
\begin{abstract}
Visual Language Models have demonstrated remarkable capabilities across tasks, including visual question answering and image captioning. However, most models rely on text-based instructions, limiting their effectiveness in human-machine interactions. Moreover, the quality of language models depends on reasoning and prompting techniques, such as COT, which remain underexplored when using speech instructions. To address these challenges, we propose \textbf{SilVar}, a novel end-to-end multimodal model that uses speech instructions for reasoning in visual question answering. In addition, we investigate reasoning techniques with levels including conversational, simple, and complex speech instruction. \textbf{SilVar} is built upon CLIP, Whisper, and LLaMA 3.1-8B, enabling intuitive interactions by allowing users to provide verbal or text instructions. To this end, we introduce a dataset designed to challenge models with speech-based reasoning tasks for object localization. This dataset enhances the model’s ability to process and explain visual scenes from spoken input, moving beyond object recognition to reasoning-based interactions. The experiments show that \textbf{SilVar} achieves SOTA performance on the MMMU and ScienceQA benchmarks despite the challenge of speech-based instructions. We believe \textbf{SilVar} will inspire next-generation multimodal reasoning models, toward expert artificial general intelligence. Our code and dataset are available here.
\end{abstract}

% \def\thefootnote{$\$$}\footnotetext{This work is supported by National Science Foundation (NSF) Grant ${\#}$2138206.}

%%%%%%%%% BODY TEXT
\section{Introduction}
\label{sec:intro}
Visual Language Models (VLMs) have gained significant attention due to their capacity to bridge the gap between visual and textual modalities, facilitating more intuitive interactions between humans and machines. These models are valuable in tasks like visual question answering (VQA), which may involve yes/no answers, multiple-choice questions, or even generating image descriptions. With advances in deep learning, VLMs can now effectively handle visual scenes and provide meaningful textual outputs that explain or describe those scenes in natural language.

Recent advancements in VLMs, such as CLIP \cite{radford2021learning}, have enabled Large Language Models (LLMs) to process images and text simultaneously \cite{ranasinghe2023language, alayrac2022flamingo, awadalla2023openflamingo}. For example, Flamingo is a VLM that can handle both modalities and excels at tasks like image captioning and VQA \cite{alayrac2022flamingo}. Similarly, BLIP-2 connects a visual encoder to LLMs using a querying transformer, creating a more efficient multimodal model \cite{li2023blip}. There are several VLM models, such as LLava \cite{NEURIPS2023_6dcf277e}, LocVLM \cite{ranasinghe2024learning}, and LISA \cite{lai2024lisa}, that further improve model predictions through reasoning and prompting techniques.

Despite the significant advancements in VLMs, most of them currently support only text-based interactions, limiting their application to scenarios where text input is inconvenient or unavailable. In addition, while reasoning and prompting techniques for LLMs have been explored, these techniques for speech-based instruction remain largely underexplored. Recently, models like GPT-4o \cite{openai2024} have enabled interaction with LLMs through speech, significantly enhancing the user experience compared to traditional text-based interactions. However, there is still a lack of exploration in the open-source community on building such speech interaction models based on foundation models.

To enable speech interaction with LLMs, speech instruction models such as Qwen2-Audio \cite{chu2024qwen2}, SALMONN \cite{tang2023salmonn}, and Llama-Omni \cite{fang2024llama} have been developed to process speech instead of text-based instructions. Although these models enable speech instruction, they are not capable of understanding both images and text simultaneously. Inspired by VLMs and ARS models, we propose \textbf{SilVar}, a multimodal model that can understand both images and audio or images and text, at the same time. In addition, we investigate reasoning techniques for speech instructions on image description and object localization. To this end, we further provide a dataset for speech instruction. The key contributions of our paper are summarized as follows:

\begin{itemize}
    \item We propose a multimodal model with speech instruction for text generation and object localization.
    \item Speech reasoning -- we investigate the effect of reasoning for speech instruction, which includes conversation level, simple reasoning, and complex reasoning.
    \item We propose a training pipeline and publicly release the reasoning speech instruction dataset.
\end{itemize}

\section{Related Work}
\label{sec.relatedwork}

With the advent of LLMs such as GPT-3 \cite{brown2020language} and GPT-4 \cite{achiam2023gpt}, as well as the rise of open-source models like the Llama family \cite{touvron2023llama, touvron2023llama2, dubey2024llama} and Vicuna \cite{zheng2023judging}, these models have paved the way for the development of VLMs. VLMs extend the capabilities of LLMs by enabling them not only to understand text input but also to learn from visual information \cite{li2023blip}. This advancement has significantly accelerated the progress of multimodal models in recent years, particularly in integrating vision and language or speech and language. Pioneering work, such as CLIP \cite{radford2021learning} and ALIGN \cite{jia2021scaling}, established a framework for combining visual models with language processing. Building on this foundation, models like Flamingo \cite{alayrac2022flamingo}, BLIP \cite{liu2024improved}, MiniGPT-v2 \cite{chen2023minigpt}, MiniGPT-4 \cite{zhu2023minigpt}, and LLava \cite{NEURIPS2023_6dcf277e} have demonstrated significant advancements in tasks such as visual question answering and image captioning. The application of VLM models has further extended to object detection, segmentation, and reasoning-based localization \cite{NEURIPS2023_6dcf277e, lai2024lisa, zhu2023minigpt, wang2024visionllm, ranasinghe2024learning}. While earlier models explored large architectures with several billion parameters, the recent trend is to develop smaller yet high-performance models or to investigate prompting techniques that make smaller models more robust \cite{wei2021finetuned, dubey2024llama, li2022competition}.

In parallel, speech recognition has emerged as a crucial area of research, particularly with the development of automatic speech recognition (ASR) systems such as Whisper \cite{radford2023robust} and Wav2Vec \cite{baevski2020wav2vec}. In addition, there are lot of work that has been done on speech-related tasks, including speech-to-text translation, speech emotion recognition (SER), and vocal sound classification (VSC) \cite{tang2023salmonn, wang2023blsp, ao2021speecht5}. Beyond speech generation, recent models have been investigated to enhance emotion and voice interactions, such as AudioPaLM \cite{rubenstein2023audiopalm} and LauraGPT \cite{du2023lauragpt}, fostering more natural communication. Innovations such as VALL-E \cite{wang2023neural} and MusicGen \cite{copet2024simple} further illustrate how audio generation can enrich text-based interactions. Furthermore, the challenge of low-resource conversational telephony speech corpora has been investigated using unsupervised learning and fine-tuning techniques of large pre-trained models \cite{vieting2023efficient}.

The integration of ASR with language models has led to the development of multimodal models \cite{chu2024qwen2, xie2024mini}. For instance, SpeechGPT \cite{zhang2023speechgpt} allows users to engage with large language models using speech. Additionally, HuggingGPT \cite{shen2024hugginggpt} enhances this interaction by discretizing speech into tokens and expanding the LLM’s vocabulary to accommodate speech inputs. Furthermore, the study in \cite{adedeji2024sound} demonstrated that LLMs have the potential to improve the accuracy of ASR systems, particularly in medical transcription.

While VLMs have made great strides in integrating vision and text, the addition of audio and ASR to these systems has paved the way for even more comprehensive multimodal models, allowing for richer and more dynamic interactions across different modalities. With the development of GPT-4o \cite{openai2024}, users can interact with large language models in real time, enhancing the user experience compared to traditional text-based interactions. However, this model is not open-sourced, which limits the ability to finetune it further. As a result, there is a lack of open-sourced models that enable speech-based interaction with VLMs. Inspired by VLMs and ASR models, we propose \textbf{SilVar}, a novel multimodal model that addresses these challenges by incorporating speech instructions directly into the reasoning process, creating a more efficient interaction model for image interpretation and object localization tasks. Beyond that, we also provide a comprehensive pipeline for building multimodal models by leveraging existing open-source foundation models, which may inspire future research into developing next-generation multimodal reasoning models.

Although LLMs play a crucial role in response generation, prompting techniques have proven equally important for reasoning tasks such as question answering \cite{NEURIPS2023_6dcf277e, lai2024lisa}. In fact, there is a strong and evident connection between the effectiveness of reasoning methods or prompting techniques and the overall performance of LLMs \cite{wei2022chain, yao2024tree}. In the context of ASR analysis, the ability of LLMs to handle complex tasks like diarization and error correction has been explored in \cite{adedeji2024sound}, where techniques such as zero-shot prompting and COT prompting were shown to influence results. Therefore, our study further explores the potential of various reasoning methods for speech instruction, including zero-shot prompting and COT reasoning on LLMs’ responses.

With the rapid development of LLMs, VLMs, and instruction-tuning techniques, several multimodal benchmarks aim to evaluate models on various aspects related to human-like skills \cite{hendryckstest2021, hendrycks2021ethics, yue2024mmmu, goyal2017making, lu2022learn}. Earlier work focused on simple question-answering or visual question-answering tasks, which, while beneficial, is insufficient for evaluating a model's understanding and reasoning capabilities \cite{antol2015vqa, marino2019ok}. Therefore, numerous datasets and benchmarks have been proposed to address this limitation, such as GAIA \cite{mialon2023gaia}, ScienceQA \cite{lu2022learn}, MMMU \cite{yue2024mmmu}, MMBench \cite{liu2025mmbench}, MMVet \cite{yu2023mm}, SEED \cite{li2023seed}, and LLaVA \cite{NEURIPS2023_6dcf277e}. Although these benchmarks are valuable for evaluating the performance of LLMs and VLMs, they are limited to text-image pair samples. Since speech instruction for VLM models is a new task and, to our best knowledge, no dataset currently exists for this purpose, we provide a new dataset named SilVar, which consists of text, images, and speech instructions along with the text generation. Additionally, we convert text into audio from existing reasoning datasets and benchmarks, including MMMU, LISA, and ScienceQA, for our training purposes.

\section{GPT-assisted Data Generation}
\label{sec.data}

% \subsection{SilVar Dataset Design}

With the development of multimodal models, there has been a surge in datasets that support model training, such as Flickr30K \cite{young2014image}, Visual Genome \cite{Krishna2016VisualGC}, and MovieQA \cite{tapaswi2016movieqa}. However, these datasets are limited to tasks like automatic image description, image or video captioning, and simple visual question-answering. To explore the understanding and explainability of multimodal models, more intricate datasets such as LAION \cite{schuhmann2022laion}, SEED \cite{li2023seed}, and LLaVA \cite{NEURIPS2023_6dcf277e} have been created, enabling LLMs to generate detailed responses. Despite this progress, the available data are insufficient for guiding LLMs in querying and responding to users' input, particularly for tasks requiring complex instructions. As a result, techniques like hard prompting \cite{wen2024hard} or prompt engineering \cite{wei2022chain, yao2024tree} have been proposed. For dataset, LLaVA \cite{NEURIPS2023_6dcf277e} is one of the recent datasets that use strong prompting baselines. However, this type of data is a text-based instruction and not well-suited for speech instruction, especially in the context of human-machine interaction. To address this, we propose a unique speech instruction dataset that emphasizes natural conversation.

\begin{figure*}[ht]
    \centering
    \includegraphics[width=1\linewidth]{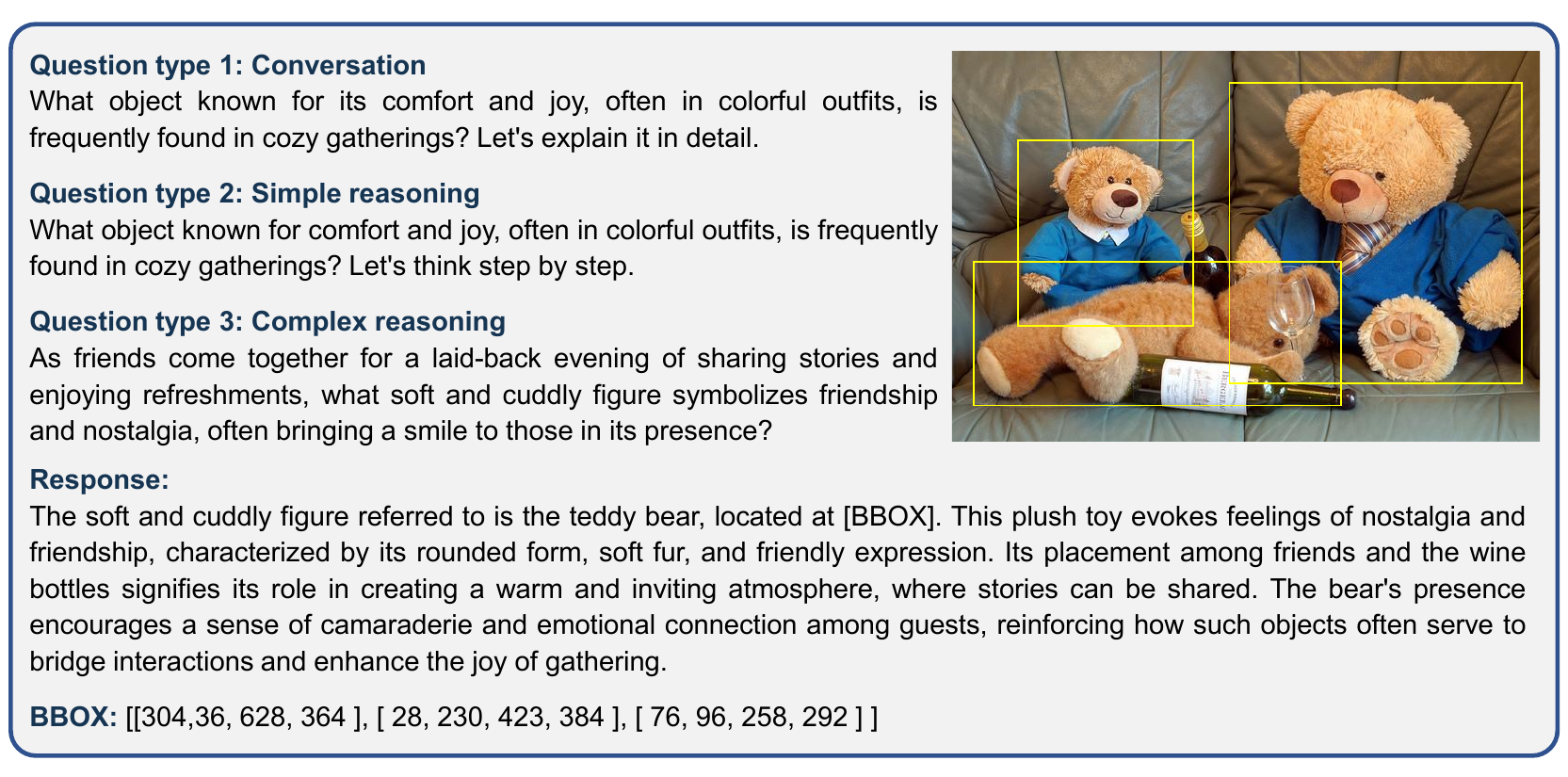}
    \caption{An example from our SilVar-bench dataset, focusing on reasoning speech instructions with different types: conversation, simple reasoning, and complex reasoning. The detected objects are highlighted in yellow bounding boxes. The dataset not only focuses on reasoning instructions but also generates visual explanations, enhancing spatial understanding and interpretability.}
    \label{fig.example_dataset}
\end{figure*}

Inspired by the success of recent GPT models in text-based tasks and GPT-assisted datasets \cite{NEURIPS2023_6dcf277e}, we developed our dataset with GPT-4 \cite{openai2024gpt4} assistance, as shown in \ref{fig.example_dataset}. Notably, we only used text as input for GPT-4 to generate different types of questions. To ensure the model can effectively perform reasoning-based object localization and generate coherent responses from speech instructions, we designed the dataset according to the following criteria:

% \begin{tcolorbox}
% [colback=black!5!white,colframe=black!100!white,title=, left=0mm,right=0mm]
\begin{itemize} 
    \item \textbf{Human-machine conversation}: The dataset is designed to reflect natural human-machine conversations, enabling the agent to interpret and respond to verbal instructions in a conversational context. 
    \item \textbf{Reasoning instructions and responses}: Unlike traditional datasets focused on simple object recognition, our dataset contains reasoning instructions and explanations behind the responses. This is crucial for enhancing the model's ability to explain not just what the object is, but also why it is located in a particular place. 
    \item \textbf{Detailed descriptions}: The dataset includes both simple questions and complex reasoning scenarios, requiring the model to provide detailed descriptions of visual scenes. 
\end{itemize}
% \end{tcolorbox}

The SilVar dataset was constructed using 998 images randomly selected from the COCO 2014 dataset \cite{lin2014microsoft}, which offers a diverse range of real-world scenes and object types. Each image is accompanied by two tasks: speech-based detailed description and complex reasoning. The dataset was generated using GPT-4, guided by a carefully designed prompt structure. For each image, GPT-4 generated three questions targeting the same object without directly naming it, focusing instead on creating scenarios that implied the object’s role and importance. Each answer was required to provide a comprehensive explanation covering:
% \begin{tcolorbox}
% [colback=black!5!white,colframe=black!10!black,title=, left=0mm,right=0mm]
\begin{itemize}
    \item \textbf{Object’s characteristics}: A description of the object's features, shape, and functionality.
    \item \textbf{Background context}: An explanation of how the object is relevant to the broader environment depicted in the image.
    \item \textbf{Interaction with surroundings}: An analysis of how the object interacts with other elements in the scene, such as people, activities, or other objects.
\end{itemize}
% \end{tcolorbox}

Following the generation of questions and answers, we implemented a rigorous human verification process to ensure the quality and coherence of the dataset. Human reviewers verified that each set of questions consistently referred to the same object and that the answers offered detailed reasoning, encompassing the object’s characteristics, background, and interactions. In addition, bounding boxes were manually labeled using Roboflow \cite{dwyer2024roboflow} to accurately pinpoint the specific objects being referred to in each image. Given the qualified dataset, we convert text to speech using Google Cloud APIs with over 50 different voices. By carefully generating data with human verification and manual bounding box labeling, SilVar-Bench aimed to create a robust evaluation framework. Its purpose is to challenge models not only in terms of object recognition but also in their ability to generate detailed, context-specific reasoning responses. This dataset is particularly suited for advancing multimodal models that require both spatial understanding and detailed reasoning in speech-driven interactions.

\begin{table}[h]
    \centering
    \begin{tabular}{llcc}
        \toprule
        \textbf{Dataset} & \textbf{Train} & \textbf{Validation} & \textbf{Test} \\ \midrule
        ScienceQA & 6,218 & 2,097 & 2,017\\ 
        MMMU & 150 & 900 & 10,500  \\ 
        LISA & 239  & 200 & 779  \\ 
        SilVar & 300 & - & 700\\ 
        \bottomrule
    \end{tabular}
    \caption{Datasets used for training and testing SilVar, including sample distributions across the training, validation, and test sets.}
    \label{tab:dataset}
\end{table}

% \subsection{Data Collection}
In addition to our dataset, we also leverage available text-based reasoning datasets for pretraining purposes, including MMMU \cite{yue2024mmmu}, LISA \cite{lai2024lisa}, and ScienceQA \cite{lu2022learn}, as these datasets are designed for reasoning localization and description. The MMMU dataset consists of 11,500 samples covering 30 subjects and 183 subfields, ranging from mathematics to art, business, and medicine, as shown in \ref{tab:dataset}. This dataset is highly suitable for training models aimed at expert-level AGI reasoning, as it contains diverse and complex tasks that require both textual understanding and visual perception, including diagrams, tables, and other intricate image types. For our study, we curate a subset consisting of 132 samples from the training set and 775 samples from the validation set. The primary goal of our preprocessing is to adapt the dataset for speech instruction models, ensuring that the samples can be effectively handled in a spoken format. The LISA dataset includes 239 training samples, 200 validation samples, and 779 test samples; however, we only use the training set for our purposes. Similarly, the ScienceQA dataset consists of 26 topics, 127 categories, and 379 skills, covering a wide range of domains. For ScienceQA, we filter out samples that do not contain images and only use the available image-text pair samples. To make the input clearer and more vocalizable for speech generation, our preprocessing involves the following key steps:

% \begin{tcolorbox}
% [colback=black!5!white,colframe=black!100!white,title=, left=0mm,right=0mm, top=1mm, bottom=1mm]
\begin{itemize}
    \item \textbf{Handling special characters}: We converted complex symbols like LaTeX and non-standard characters into formats suitable for speech instruction, enabling correct processing and vocalization.
    \item \textbf{Punctuation and text normalization}: We standardized punctuation in the text to ensure that it would generate smooth, natural speech outputs, enhancing the clarity and coherence of the spoken content.
\end{itemize}
% \end{tcolorbox}

\section{Speech Instruction Tuning}
\label{sec.SilVar}

\subsection{Architecture}
\label{sec.architecture}

\begin{figure*}[h]
    \centering
    \includegraphics[width=0.98\linewidth]{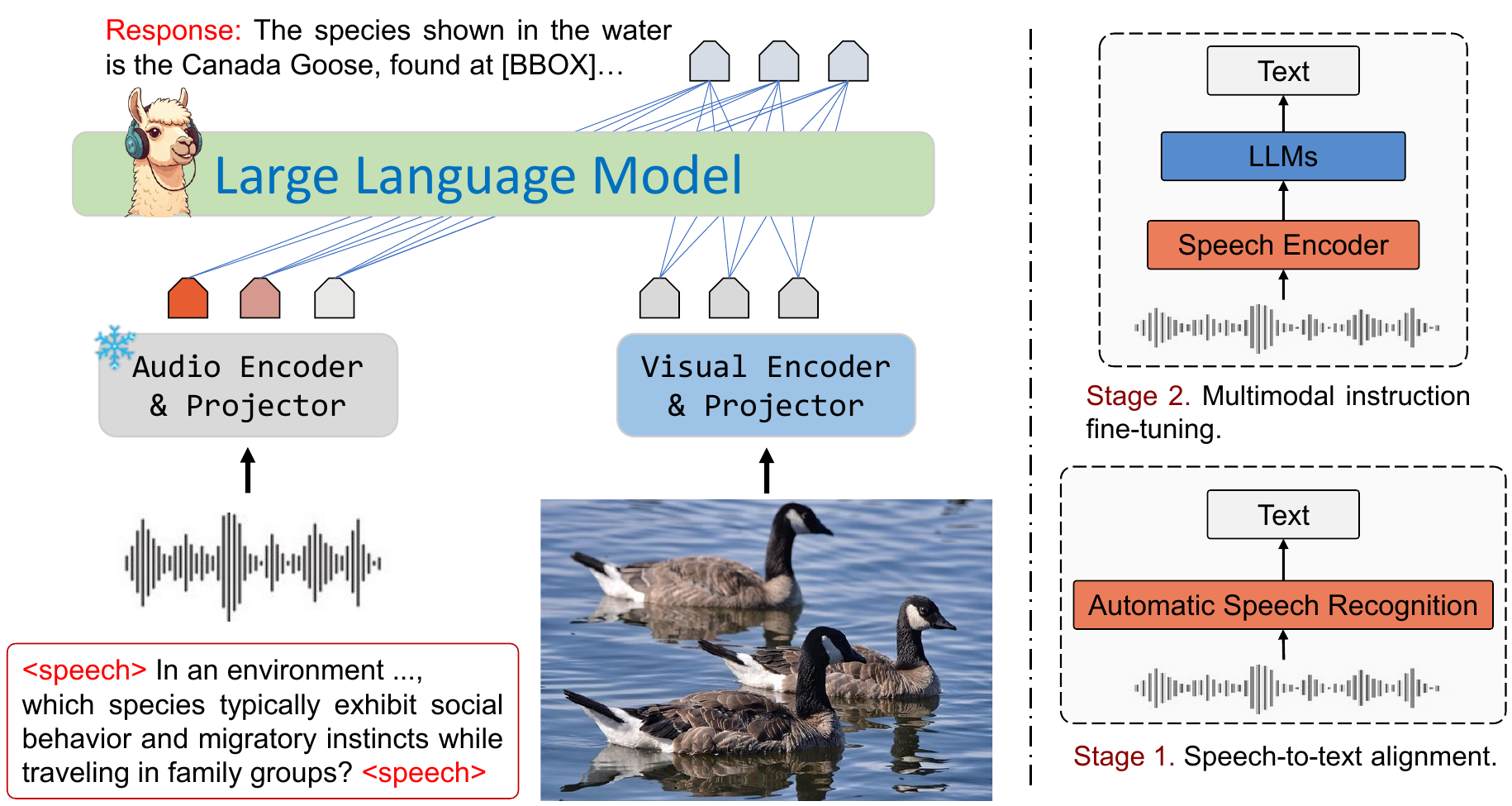}
    \caption{Illustration of the SilVar model architecture, integrating visual and audio instruction for reasoning text generation and object localization. The model comprises four key components: an audio encoder for extracting features from speech, a visual encoder for processing images, a projector for feature transformation, and an LLM that processes information across modalities to generate coherent responses.}
    \label{fig:SilVar}
\end{figure*}

SilVar is a multimodal model designed for image interpretation and object localization based on various input modalities, including speech, text, and images. This multimodal approach enables SilVar to understand and respond comprehensively to a wide range of inputs and prompts. The model architecture is illustrated in \ref{fig:SilVar}, consisting of several key components: an audio encoder, a visual (image) encoder, a projector, and a language model.

\textbf{Audio Encoder}: This module extracts relevant features from audio data, providing critical context for understanding spoken content rather than text-based instructions. We use Whisper, a model widely utilized in speech recognition and foundational models \cite{radford2023robust}. Whisper is trained on thousands of hours of data, showing strong semantic alignment in speech recognition. Specifically, we use the tiny Whisper variant, which has only 39 million parameters, making it lightweight yet effective in speech recognition tasks \cite{moor2023foundation}. Given an input audio $A$ with a maximum length of 1500, consisting of speech instructions that may contain questions about an image, the Whisper encoder extracts meaningful features and embeds them into a 768-dimensional feature vector. These features are then passed through an audio projector, which contains a Linear layer to match the LLM’s input ($\mathbb{R}^{4096}$). These features are later concatenated with image embeddings of the same size. For the audio adapter, we later investigate the effects of the different types of neural networks including MLP and Transformer layers.

\textbf{Visual Encoder}: Similar to the audio encoder, this module processes input images to extract meaningful features. For this purpose, we use CLIP with pre-trained weights, specifically the ViT-B/32 model, as the visual encoder. %, which converts incoming images to a feature presentation. 
The CLIP is wisely trained on over 400 million (image, text) pairs collected from the internet, making it capable of understanding most images and descriptions with a certain level \cite{radford2021learning}. For each image input $X \in \mathbb{R}^{H \times W \times C}$, where $H$, $W$, and $C = 3$ denote the image's height, width, and color channels, respectively, the visual encoder processes the input image by resizing it to a standardized shape of $224 \times 224$ pixels. Subsequently, CLIP generates a feature representation of the input image, which is a sequence of 768-dimensional visual tokens, representing the content. 

To integrate this visual representation with the language model, we project the 768-dimensional visual tokens to a higher-dimensional space ($\mathbb{R}^{4096}$) through a visual adapter. Inspired by MiniGPT-v2 \cite{zhu2023minigpt}, we designed the adapter with two Linear layers and the GELU activation function \cite{hendrycks2016gaussian}, which serves as an essential bridge between the vision backbone and the language model. After that, the output of the visual adapter is concatenated with audio embeddings produced by the audio adapter, creating a multimodal feature representation. During the model development process, we observed that by concatenating multiple encoders, not limited by either pair of images and text or images and audio, we can create a robust, versatile model that is capable of handling diverse input types.

\textbf{Large Language Model}: At the core of SilVar is a language model, responsible for generating text and bounding boxes by processing information from the audio and visual encoders. To this end, we decided to use LLama 3.1-8B \cite{dubey2024llama}, a novel open-source model, which was trained on a diverse range of text-related tasks. LLama serves as the foundational model, allowing us to effectively transfer its pre-trained knowledge into domain-specific tasks such as localization where understanding both verbal and visual inputs. By combining visual and audio tokens, we provide a diverse set of token embeddings for the language model, requiring it to process and generate a final representation of textual tokens. After processing these tokens, the language model outputs tokens that are untokenized to form natural language text, which is the model's response to the input prompt. The integration of these components enables SilVar to generate human-quality text responses by leveraging the complementary strengths of each modality, making it a robust system for multimodal instruction and interaction.

\subsection{Training}
\label{sec.training}
We propose a two-step training process for speech instruction: (1) speech-to-text alignment and (2) LLM training response, as shown in \ref{fig:SilVar}. Since speech plays an important role as an instructional modality in our model, we start training a speech-to-text system to align audio with text in the domain of reasoning text generation. We use the tiny Whisper model for speech recognition due to its efficiency in real-time speech processing. As mentioned in Section \ref{sec.architecture}, with the purpose of developing a foundation model that can understand and handle numerous tasks, we train the model on the ScienceQA and MMMU datasets using speech instructions, as these datasets are reasoning benchmarks for human-level understanding and explanation \cite{lu2022learn, yue2024mmmu}. This step is crucial for enabling SilVar to process spoken instructions effectively. Furthermore, we aim to specialize the model for the task of object localization; therefore, we further train the tiny Whisper on the LISA dataset and our benchmark. The data type of each dataset and the stages used in the training process are shown in \ref{tab:datasets_and_stage}.

\begin{table}[ht]
    \begin{tabular}{cccc}
        \toprule
        \textbf{Dataset} & \textbf{Data type} & \textbf{Stage 1} & \textbf{Stage 2}  \\ \midrule
        ScienceQA & Text, image & x & x  \\ 
        MMMU & Text, image & x & x \\ 
        LISA & Text, image & x & x \\ 
        SilVar & Text, image, speech & x & x  \\ 
        \bottomrule
    \end{tabular}
    \caption{Summary of datasets and corresponding stages used in our training process.}
    \label{tab:datasets_and_stage}
\end{table}

In stage 2, we leverage the pre-trained weights from stage 1 to train our model for the visual question-answering task, using direct audio input from the audio encoder for reasoning. Similar to stage 1, we use the ScienceQA, MMMU, LISA, and SilVar datasets for training text generation. In our experiments, we use AdamW optimizer to update the parameters \cite{loshchilov2019}, accompanied by a learning rate scheduler with a linear warmup followed by a cosine decay. The learning rate begins at 1e-5, with both the minimum and warmup rates also set to 1e-5. The warmup spans 1000 steps, and we apply a weight decay of 0.05. The model is trained for up to 20 epochs, with each epoch including 1722 iterations and a batch size of 4, utilizing 2 workers. Training is conducted on a computing system equipped with four A100 GPUs and requires around twenty-two hours.

% Our model training incorporates low-rank adaptation (LoRA) \cite{hu2021lora}, a parameter-efficient fine-tuning approach, where we load the model in 8-bit precision and initialize the $W_q$ and $W_v$ parameters using the original weights. The rank parameter, $r$, is set to 64, and we continue fine-tuning these parameters through lower-rank adaptation.

\section{Experiment and Result}
\label{sec.experiment}

To evaluate the performance of the SilVar model, we test it with different abilities: reasoning for text generation, localization, chatting capability, and reasoning ability on AGI expert benchmarks. In addition, we also compare its performance when we use different input modalities either image-text pairs or image-speech pairs.

\subsection{Speech Instruction \& Text Instruction}
% \textcolor{red}{HERE!!!! Compare speech instruction and text instruction (cider, blue, meteor, rouge, and bousing box accuracy)}

To assess the effect of different instructional modalities, we compare SilVar's performance using speech-based and text-based instructions across various reasoning and conversational tasks. This comparison includes evaluating SilVar’s ability to understand and respond to complex and simple reasoning instructions as well as conversational prompts in both speech and text formats. It is important to note that we train SilVar only with complex reasoning and use other reasoning techniques for evaluation. For this evaluation, we apply key metrics for text generation validity, including CIDEr, BLEU, METEOR, and ROUGE. Additionally, for localization capability, we use bounding box accuracy with an intersection over union (IoU) threshold of 0.5. Together, these metrics provide a comprehensive view of both object detection and text generation performance.

\begin{table*}[h]
    \centering
    \begin{tabular}{lcccccc}
        \toprule
        \textbf{Instruction Type} & \textbf{ROUGE-1 } & \textbf{BLEU-1 } & \textbf{METEOR } & \textbf{CIDEr} & \textbf{Accuracy (IoU = 0.5)} \\ 
        \midrule
        Complex reasoning (speech) & \textcolor{cyan}{34.48} & \textcolor{cyan}{37.14} & \textcolor{cyan}{26.57} & \textcolor{cyan}{0.04} & \textcolor{cyan}{23.31} \\
        Simple reasoning (speech) & 33.31 & 34.61 & 24.14 & 0.04 & 22.52   \\
        Conversation (speech) & 33.32 & 34.75 & 24.07 & 0.03  & 21.11 \\
        \midrule
        Complex reasoning (text) & \textcolor{blue}{36.51} & \textcolor{blue}{38.40} & \textcolor{blue}{27.86} & \textcolor{blue}{0.07} & \textcolor{blue}{26.32}   \\
        Simple reasoning (text) & 34.86 & 35.02 & 26.23 & 0.06 & 25.97  \\
        Conversation (text) & 34.64 & 35.00 & 26.42 & 0.06 & 24.56   \\
        \bottomrule
    \end{tabular}
    \caption{Performance of SilVar model on various instructional types (conversational, simple, and complex) using speech- and text-based modalities. The highlighted values in blue represent the highest scores achieved for each metric in text-based complex reasoning, while values in cyan highlight the highest scores for speech-based complex reasoning.}
    \label{tab:instruction_metrics_SilVar}
\end{table*}

\ref{tab:instruction_metrics_SilVar} shows SilVar’s performance across instruction types, between the text-based reasoning performance compared to speech-based instructions. Remarkably, text-based instructions consistently score higher across most metrics, with complex reasoning (text) achieving top scores: ROUGE-1 (36.51), BLEU-1 (38.40), METEOR (27.86), and CIDEr (0.07). In terms of object detection, the accuracy at IoU = 0.5 is 26.32\% for text-based complex reasoning, indicating more precise localization when instructions are text-based. For speech-based tasks, the complex reasoning technique achieves the highest scores, though it remains slightly lower than text-based equivalents, likely due to the inherent challenge of accurately interpreting and aligning spoken language with visual information. These results highlight the potential of the SilVar model with speech instructions as a promising alternative for interpreting spoken language effectively, which could broaden accessibility and user engagement in multimodal applications.

\subsection{MMMU-bench}
MMMU is a multimodal dataset with tasks that demand college-level knowledge and deliberate reasoning, which enables the test of the performance of models in terms of expert-level perception and reasoning. The dataset includes 30 subjects across various disciplines, such as art, science, health \& medicine, and engineering. In particular, we compare our SilVar model to state-of-the-art (SOTA) models with a similar or greater number of parameters. In addition, since the test dataset is significantly larger than the validation dataset, and SOTA models achieve similar performance on both test and validation datasets, we intentionally evaluate our model only on the validation set.

\begin{table}[ht]
\centering
\small
\begin{tabular}{lccc}
\toprule
\textbf{Model}    & \textbf{Instruction}    & \textbf{Val} & \textbf{Test} \\ \midrule
Adept Fuyu-8B  \cite{fuyu-8b}   & Text    & 27.9	& 27.4                 \\ 
OpenFlamingo2-9B \cite{awadalla2023openflamingo}  & Text    &  28.7	& 26.3                   \\
MiniGPT4-Vicuna-13B \cite{zhu2023minigpt}	& Text & 26.8	& 27.6          \\ 
LLaMA-Adapter2-7B\cite{zhang2023llama} & Text& 29.8	& 27.7 \\
LLaVA-1.5-13B \cite{liu2024improved} & Text &	36.4 & 33.6 \\
Qwen-VL-7B-Chat \cite{bai2023qwen} & Text &	35.9 &	32.9 \\
\midrule
% \rowcolor{gray!20} \multicolumn{4}{l}{\textbf{Result of our model}} \\ 
\rowcolor{gray!20} SilVar & Text & 31.8 &  -  \\
\rowcolor{gray!20} SilVar & Speech & 30.2 &  -  \\
\rowcolor{gray!20} SilVar-end-to-end & Speech & 30.4 &  -  \\

\bottomrule
\end{tabular}
\caption{Performance of SilVar and other models on the MMMU benchmark.}
\label{tab:mmmu_benchmark}
\end{table}

Unlike most models, SilVar can process both text and speech inputs, allowing us to evaluate its unique capability against models designed primarily for text instructions. As shown in \ref{tab:mmmu_benchmark}, SilVar's performance is comparative or surpasses many baseline models, particularly in the text-based instruction setting, demonstrating its robustness in complex reasoning tasks. In particular, SilVar achieves a validation score of 31.8 with text-based instructions, making it competitive among SOTA models such as LLaVA-1.5 and Qwen-VL-7B-Chat, which achieves 36.4 and 35.9 scores on the MMMU benchmark, respectively. Additionally, with speech-based instructions, SilVar maintains strong performance with a validation score of 30.2, demonstrating its versatility in handling multimodal inputs effectively. To further investigate the model's capabilities, we train it end-to-end, aiming to enhance its overall performance. However, the model's performance is slightly improved, with a score of 30.4.

\subsection{ScienceQA Benchmark}

Similar to the MMMU benchmark, we also evaluate our model on the ScieneQA benchmark. In this benchmark, we compare the SilVar to the SOTA models with a similar approximate number of parameters, and we use only speech-based instructions instead of text-based instructions. As shown in \ref{tab:scienceqa}, SilVar achieves an average accuracy of 63.21\%, demonstrating its effectiveness in processing across multimodal input types. Compared to models such as LLaMA-Adapter, Chat-UniVi, and LaVIN, which achieved higher overall scores by leveraging text instructions, SilVar shows promising performance in scenarios where only speech instructions are available or preferable.

\begin{table}
\centering
% \resizebox{\textwidth}{!}{%
\begin{tabular}{lcccccccccc}
\toprule
\textbf{Model} & \textbf{Instruction} & \textbf{Average Score}\\ \midrule

LLaMA-Adapter \cite{zhang2023llama} & Text  & 85.19 \\ 
Chat-UniVi (7B) \cite{jin2024chat}& Text &  88.78 \\
% LaVIN-7B \cite{luo2024cheap} & Text & 89.25 & 94.94 & 85.24 & 88.51 & 87.46 & 88.08 & 90.16 & 88.07 & 89.41 \\
LaVIN-13B \cite{luo2024cheap} & Text  & 90.83 \\
LLaVA-13B \cite{NEURIPS2023_6dcf277e} & Text  & 90.92 \\
% PILL (LLaMA-7B) \cite{luo2024cheap} & Text & 90.36	 & 95.84 & 89.27 & 89.39 & 88.65 & 91.71 & 92.11 & 89.65 & 91.23 \\ 
LLaVA-13B \cite{yang2023mm} & Text & 47.74 \\
MiniGPT-4 \cite{zheng2023ddcot} & Text &  47.71 \\
LLaVA-7B \cite{yang2023mm} & Text & 41.1 \\
OpenFlamingo \cite{yang2023mm} & Text &  39.27 \\

\midrule

% \rowcolor{gray!20} \multicolumn{11}{l}{\textbf{Result of our model}} \\ 
\rowcolor{gray!20} SilVar  & Speech  &  63.21 \\
\rowcolor{gray!20} SilVar-end-to-end & Speech &  63.45  \\
\bottomrule
\end{tabular}%
% }
\caption{Performance comparison of our model and SoTA models on the ScienceQA Benchmark.}
\label{tab:scienceqa}
\end{table}

\subsection{Chatbot models}

After training the model, we compare the prediction from SilVar and commercialized chatbots such as GPT-4o and Gemini 1,5 Pro. As shown in \ref{tab:chatbot}, all three models correctly identified the answer as "Pigeon." However, the closed-source models exhibited limitations in their explanations and lack of visual context. Specifically, while GPT-4o provided only the answer, Gemini 1.5 Pro attempted to explain its reasoning based on the original question but still failed to include a bounding box for the answer. In contrast, SilVar not only provided the correct answer but also explained it and included the bounding box.

\begin{table*}[ht]
\centering
\begin{tabular}{>{\arraybackslash}m{10.5cm} >{\centering\arraybackslash}m{6cm}}
\toprule
\rowcolor{gray!20} \multicolumn{2}{l}{\textbf{User input}:} \\
Amid a gathering of wildlife, particularly in urban environments where remnants of human activity are abundant, which bird, typically characterized by its adaptability and social behavior in flocks, is observed standing alone?& 
\includegraphics[width=6cm]{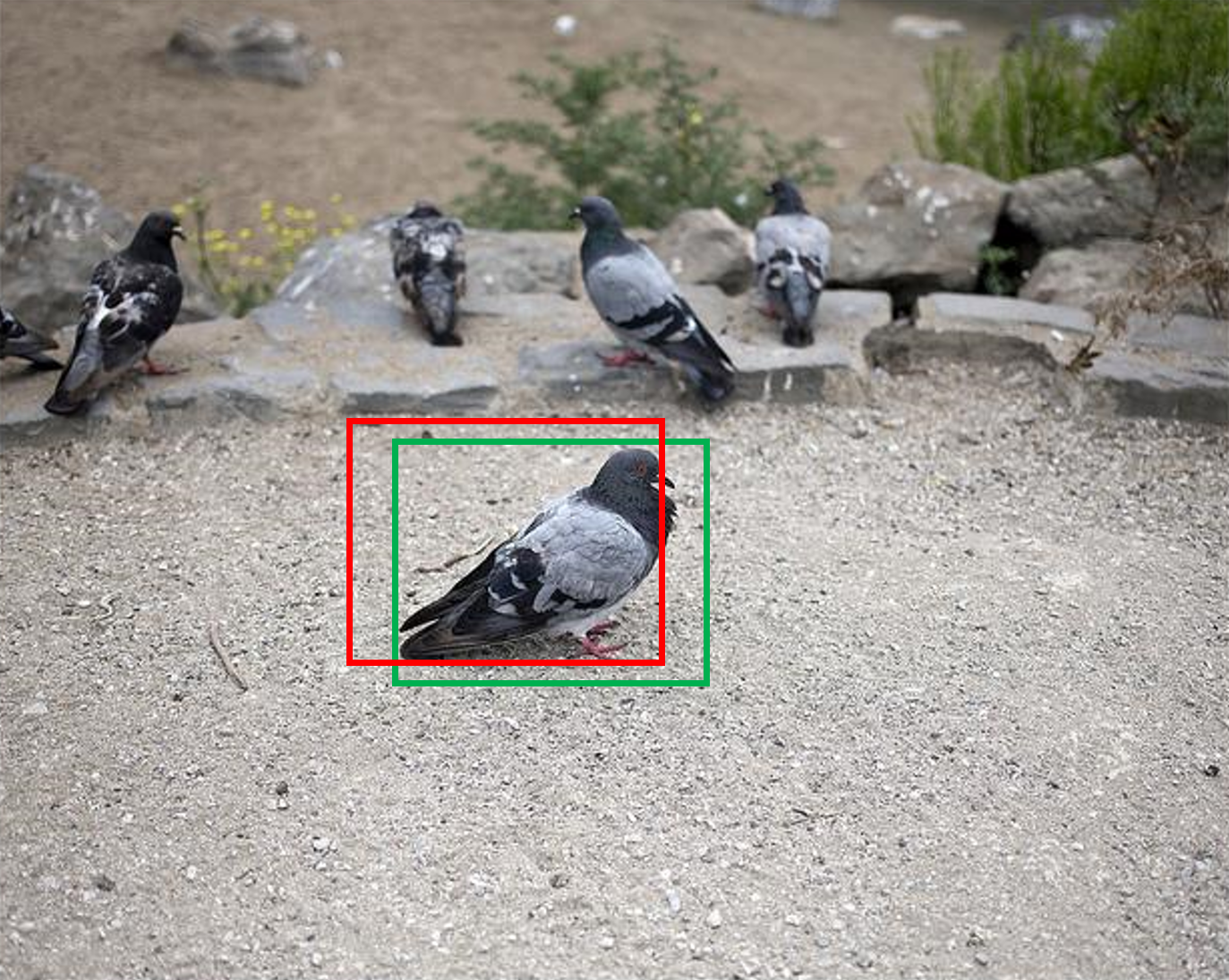} \\ 
\midrule
\rowcolor{gray!20} \multicolumn{2}{l}{\textbf{SilVar's response}: } \\ 
\multicolumn{2}{p{16cm}}{The bird mentioned is a pigeon, located at \{198, 221, 335, 355\}. This species is known for its adaptability to urban environments and its tendency to thrive near humans, often in areas with abundant food remnants.} \\ \\

\rowcolor{gray!20} \multicolumn{2}{l}{\textbf{GPT-4o's response}: } \\ 
\multicolumn{2}{p{16cm}}{Pigeon.} \\ \\

\rowcolor{gray!20} \multicolumn{2}{l}{\textbf{Gemini 1.5 Pro's response}: } \\ 
\multicolumn{2}{p{16cm}}{The image shows a group of pigeons. Pigeons are known for their adaptability and social behavior, often found in flocks, particularly in urban areas.} \\ 
\bottomrule
\end{tabular}
\caption{Example of chatting response between SilVar, GPT-4o, and Gemini 1.5 Pro. In the image, the ground truth and prediction bounding boxes are highlighted by green color and red color, respectively.}
\label{tab:chatbot}
\end{table*}

\section{Ablation study}

While the visual and audio encoders play an important role in processing image-speech pair instructions, the visual and audio adapters are also crucial for transferring information from the encoders to LLMs. In Llava \cite{liu2024improved}, linear projection and MLP-based adaptations demonstrate their effectiveness in transforming information from encoders to LLMs. The performance of models also improved by using different kinds of neural networks \cite{chen2020improved, chen2020simple}, and the choice of adapter is critical in multimodal models such as the Q-Former in BLIP-2 \cite{li2023blip} or the Perceiver Resampler in Flamingo \cite{alayrac2022flamingo}. Inspired by previous studies, we designed an MLP and a Transformer-based adapter for the audio encoder. 

\begin{table}[ht]
\centering
\small
\begin{tabular}{lccc}
\toprule
\textbf{Adapter}    & \textbf{SilVar}    & \textbf{MMMU (val)} & \textbf{ScienceQA} \\ \midrule
Linear layer  & 23.31    & 30.04	& 63.45  \\ 
MLP  & 24.04    &  31.16	& 63.41  \\
Transformer	& 24.29 & 31.05	& 63.78  \\
\bottomrule
\end{tabular}
\caption{Performance of SilVar and other models on the MMMU benchmark.}
\label{tab:ablation_study}
\end{table}

The performance of SilVar with different audio adapters is shown in \ref{tab:ablation_study}, where we train the model end-to-end using speech for instruction. It is important to note that the Transformer-based adapter takes one and a half times longer to train compared to the MLP-based adapter. The performance differences among various adapters are minimal, suggesting that by using the final layer from Whisper's encoder, there is no need for complex adapter designs to effectively transfer signals from speech to language models. In addition to neural network investigation, we explore the effect of hidden layer sizes in the MLP adapter, specifically 2816 and 5632. As a result, SilVar's performance on the MMMU dataset fluctuates with a deviation of $\pm{0.1}$.

\section{Conclusion}

In this paper, we propose \textbf{SilVar}, an end-to-end multimodal model designed to bridge the gap in human-machine interaction by enabling effective reasoning from both text and speech instructions. Unlike existing VLMs and LLMs that primarily rely on text inputs, SilVar incorporates open-sourced models, CLIP, Whisper, and LLaMA 3.1-8B, to support reasoning verbal instructions, aiming to enhance its potential for more intuitive, natural interactions. Our work explored reasoning techniques for speech instructions, from simple to complex reasoning, which have been used in current LMM and VLM research.

In addition, we developed a new dataset for challenging multimodal tasks involving speech-based reasoning for object localization, allowing SilVar to achieve comprehension beyond simple recognition. Through our experiments, SilVar demonstrates state-of-the-art performance on benchmarks including MMMU and ScienceQA, showing its strength toward expert-level AI for multimodal interactions.

\balance
{\small
\bibliographystyle{ieee_fullname}
\bibliography{egbib}
}

\end{document}